\documentclass[11pt]{article}
 
\usepackage{amssymb}
\usepackage{amsthm}
\usepackage{amsmath}
\usepackage{graphicx}

\begin{document}




\title{Computational Intelligence for Deepwater Reservoir \\Depositional Environments Interpretation}


\author{Tina Yu, Dave Wilkinson, Julian Clark and Morgan Sullivan}
\date{}
\maketitle

\begin{abstract}
Predicting oil recovery efficiency of a deepwater reservoir is a challenging task. One approach to characterize a deepwater reservoir and to predict its producibility is by analyzing its depositional information. This research proposes a deposition-based stratigraphic interpretation framework for deepwater reservoir characterization. In this framework, one critical task is the identification and labeling of the stratigraphic components in the reservoir, according to their depositional environments. This interpretation process is labor intensive and can produce different results depending on the  stratigrapher who performs the analysis. To relieve stratigrapher's workload and to produce more consistent results, we have developed a novel methodology to automate this process using various computational intelligence techniques. Using a well log data set, we demonstrate that the developed methodology and the designed workflow can produce finite state transducer models that interpret deepwater reservoir depositional environments adequately.
\end{abstract}






\section{Introduction}
The petroleum industry is increasingly moving exploration into the deepwater realm to meet the growing demand for oil and gas.  
Deepwater reservoir projects involve a large amount of financial investment; consequently, business decisions need to be made based on a clear understanding of the producibility of the reservoirs.

One key geologic characteristic that has strong impact on a reservoir's oil recovery efficiency is its depositional environment: the area in which and physical conditions under which sediments are deposited. These include sediment source and supply, depositional processes such as deposition by wind, water or ice, and location and climate, such as desert, swamp or river.  Most deepwater reservoirs are deposited in a wide range of depositional systems,  
and occur at a variety of temporal and spatial scales. Prediction of net-to-gross, continuity, architecture, and quality of these reservoirs therefore requires integration of seismic, well-log, and core data with appropriate subsurface and outcrop analogs \cite{shanmugam}. 

When conducting such a comparative analysis, it is critical that similar stratigraphic components are compared to one another. The first step in the analysis, therefore, is to  identify and label each of the stratigraphic components according to their depositional environments (see Section \ref{interpretation}). This interpretation process is labor intensive and can produce different results depending on the  stratigrapher who performs the analysis. With the goal of relieving stratigrapher's workload and to produce more consistent results, this research developed a novel methodology to automate this process using various computational intelligence techniques. In particular, we treat the interpretation problem as a language translation problem. The task is to translate a sequence of symbols from one language (reservoir well log data) into another language (the depositional labels). 

Finite state transducer (FST)  techniques have been widely applied to human language translation, e.g. text-to-speech pronunciation modeling \cite{gildea_Jurafskyd} and the parsing of web pages \cite{hsu_dung}. In this research, we applied an evolutionary computation system to generate FSTs that translate a sequence of reservoir well log data into a sequence of depositional labels
We also conducted a case study on a deepwater reservoir data set using the developed methods. The results are comparable to that produced by the stratigrapher on this data set. As the project is at the concept-proofing stage, only one data set was provided for us to model the process and to design the workflow. Encouraged by the initial success, we will continue the project and acquire new data sets from other deepwater reservoirs to validate the developed methods.

The paper is organized as follows. Section \ref{interpretation} presents the proposed deposition-based stratigraphic interpretation framework. The developed methodology for stratigraphic interpretation and the designed workflow are then presented in Section \ref{method}. Section \ref{case} gives the data set used to conduct our case study.  Following the designed workflow step by step, we give detailed explanation of how we carried out the workflow and then present their results in Section \ref{segmentation}, \ref{fuzzy}, \ref{co-evolution} and \ref{fuzzyfst}. 

In Section \ref{segmentation}, we describe the segmentation of  well log data. In Section \ref{fuzzy}, we explain the transformation of these segments into fuzzy symbols. In Section \ref{co-evolution}, we present a co-evolutionary genetic programming system to trains five classifiers. In Section \ref{fuzzyfst}, we report the training of FSTs based on the transformed fuzzy symbols and the evolved 5 classifiers. We discuss the limitations of our framework and outline our future work in Section \ref{discussions}. Finally, we conclude the paper in Section \ref{conclude}.

\section{A Stratigraphic Interpretation Framework}\label{interpretation} 
One systematic approach to identify and label stratigraphic components of deepwater reservoirs is by describing them within a hierarchical framework that is based solely on the physical attributes of the strata and is independent of the thickness and time.  In this framework, the fundamental building block of this hierarchical classification is an \textit{element}, defined as the cross-sectional characterization of the volume of sediment deposited within a single cycle of deposition and bounded by an avulsion or abandonment.  With this classification scheme, individual elements exhibit a predictable change from axis to margin in grain size, litho-facies type and architectural style. Meanwhile, since avulsion, which is the lateral shifting of a channel or lobe, controls the distribution of these characteristics,  elements can be used to understand the distribution of reservoir and non-reservoir facies.  

\begin{table*}
\caption{Summary of the lithology and GR responses of the 5 different depositional types.}
\begin{center}
{\small
\begin{tabular}{|c|c|c|}
\hline
\textbf{Lithology/Texture}
&\textbf{Gamma ray response}
& \textbf{Depositional label}
\\
\hline
\hline
 massive sandstone & Sharp based and blocky &  Channel-axis (A)  \\ \hline
 sandstone and shale & Weakly blocky  &  Channel-off-axis (OA)  \\ \hline
 mix of sandstone & &   \\ 
 inter-bedded with shale & High & Channel-margin (M) \\\hline
 shale inter-bedded  &  &   \\ 
 with sandstone & Irregular& Over-bank (OB)  \\ \hline
mass wasting & &    \\ 
(muddy or sandy) & Irregular and chaotic &Mass transport complexes (MTC)  \\ \hline
\end{tabular}}
\vspace{-0.4cm}
\label{framework}
\end{center}
\end{table*}

Two or more elements of similar grain size, litho-facies and architectural style form a \textit{complex}.  Elements within a complex are genetically related and exhibit a predictable organization and depositional trend.  A \textit{complex set} is comprised of either individual complexes of different architectural style and/or complexes of similar architectural style that exhibit depositional trends independent of one another. The description of deepwater sand-bodies utilizing this hierarchical approach provides a powerful methodology to directly compare similar stratigraphic components and greatly improve reservoir characterization and the prediction of productivity.

The elements that are of particular interest are \textit{channel}-related as they are the areas where hydrocarbon (oil and gas) deposit. For a finer characterization of a reservoir, we subdivide channel-elements into \textit{channel-axis}, \textit{channel off-axis}, and \textit{channel-margin} associations. 

Channel-axis deposits (A) are dominated by highly-amalgamated, massive sandstones deposited by high-concentration turbidity currents. They normally exhibit a sharp-based, blocky \textit{gamma ray} (GR) response. 
The channel off-axis association (OA) is composed of stacked, semi- to non-amalgamated, massive to planar-stratified sandstones with inter laminated shales. This type of deposits typically display weakly blocky to moderately serrated GR log character. 
The channel-margin deposits (M) contain a variety of litho-facies and are characterized by a hetero-lithic mixture of high- and low-concentration turbiditic sandstones interbedded with thick shales. They generally exhibit a serrated and high GR log response.  

\begin{figure}
\centering
\includegraphics[height=2.5in, width=4.5in]{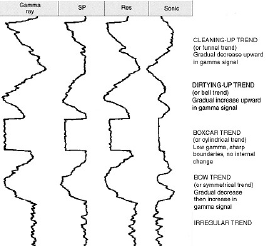}
\caption{An example of Gamma Ray (GR) well log.}
\label{gammaray}
\end{figure}

Two other element types that are non-channel and need to be identified and separated from channel elements are \textit{overbank} and \textit{mass transport complex}. Overbank deposits (OB) are dominated by shale and interbedded with thin sandstones which display an irregular character, lacking a distinct GR log trend.  Mass transport complexes (MTC), on the other hand, consist of aggregated components dominated by mass transport.  Mass wasting of basin margins and the influx of large quantities of resedimented material may occur at any time as a basin fills.  Depending on their source, these complexes can either be very muddy or very sandy, but all tend to be internally chaotic.  Due to the lithologic variability of MTC, it is difficult to uniformity characterize their log response, but commonly they display an irregular, chaotic character with an elevated GR response. 

Table \ref{framework} summarizes the lithology and the GR responses of the 5 deposition types (A, OA, M, OB and MTC). According to this classification scheme, the five deposition types have different responses to GR. It is natural for a stratigrapher to use GR well log data (see Figure \ref{gammaray}) as an indicator to classify the 5 types of deposition. The following section presents the workflow we developed to build a computer system that automates this classification task, based on GR well log data. 

\section{Methodology and Workflow}\label{method}

\begin{figure}[htp]
\vspace{-4.3cm}
\hspace{3cm}
\includegraphics[width=5.0in,height=6.05in]{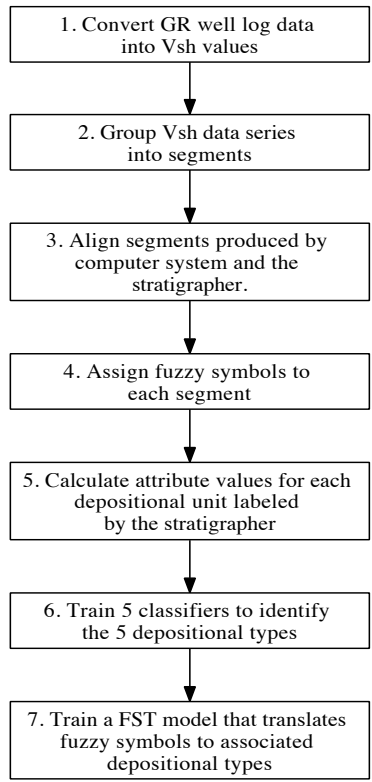}
\vspace{-1cm}
\caption{Workflow to construct a computer well-log interpretation system.}
\label{flow}
\end{figure}

We applied two computational intelligence techniques, Fuzzy Logic \cite{zadeh} and Evolutionary Computation \cite{holland} 
to train FST models that interpret GR log to identify the 5 different types of deposition. Figure \ref{flow} gives the series of steps developed to achieve that goal. 

In Step 1, GR well log data (whose range is between 0 and 150 API gamma ray units) are converted to \textit{volume of shale} (Vsh) using Equation \ref{eq:eq1}.

\scriptsize
\begin{equation}
Vsh_i= \left\{ \begin{array}{ll} 0, & GR_i < GR_{min};\\
\frac{GR_i-GR_{min}}{GR_{max}-GR_{min}}, & GR_{min}\le GR_i\le GR_{max};\\
1, & GR_i >GR_{max}.
\end{array}\right.
\label{eq:eq1}
\end{equation}
\normalsize

The purpose of this transformation is to normalize the data while maintaining the trend in the original data. 
This normalization is necessary for the later step of symbol assignment (Step 4). Note that in Equation \ref{eq:eq1}, $GR_{min}$ is the minimum GR reading in the data set and $GR_{max}$ is the maximum GR reading in the data set. 

Step 2 groups the Vsh data series into segments. The segmentation details are provided in Section \ref{segmentation}. 
This step mimics the blocking process (grouping similar consecutive GR data into a segment) that a stratigrapher carries out when performing interpretation.  By examining segmented GR data and their associated thickness, a stratigrapher can decide the depositional type of that segment. 

Segments produced by computer systems are likely to be slightly different from that blocked by humans as computers can calculate with a high degree of precision while human eyes can not. 
Normally, the differences are in the form of small edge effects. Since we value computer precision, Step 3 adjusts the boundary locations produced by the stratigrapher to align with that generated by the computer. The adjusted segments will be used in the later steps of classification rules and FST training. 

Note that this step is only performed on training data. For new GR well log, this step is skipped and the data preprocessing only consists of Step 1, 2 and 4.

The adjusted segments of Vsh data are represented as a series of numerical values, $\overline{Vsh}={\overline{s_1},\overline{s_2},\overline{s_3}\dots}$, where $\overline{s_i}$  is the average of the data values within segment $S_i$. Numerical values are continuous and less tolerant to uncertainty.  By contrast, symbols are discrete and easier for computer to manipulate and to carry out the interpretation task. We therefore simplified the numerical values using symbols. Based on our previous experiences in well log segmentation \cite{yu_wilkinson1} \cite{yu_wilkinson2}, we decided to use 4 symbols  (a, b, c, d) to represent the Vsh data. We first assign each of the 4 symbols with a different Vsh value range. Each $\overline{s_i}$ is then converted to its associated symbol. 

To enhance the expressive power of the representation, we used fuzzy, instead of crisp, symbols so that
$\overline{s_i}$ values which lay between the boundaries of two symbols can be represented by both symbols (Step 4).
The implementation details of fuzzy symbols assignment is provide in Section \ref{fuzzy}.
Using the fuzzy symbols, the next step is to 
train FST models that translate these fuzzy symbols to depositional labels.

A FST is a model that maps a string of symbols in a source language into a string of symbols in a target language. The output symbol is determined by two factors: the current input symbol and the current state. However, stratigraphic interpretation is more complicated. In addition to GR reading, a stratigrapher also considers other factors, such as the thickness of each segment and the degree of variation of neighboring segments, to give interpretation. In other words, the output symbol is decided by a model, in addition to the current input symbol and the current state. This model is a decision model that gives one of the 5 possible depositional labels as its output.

We construct the decision model using Genetic Programming (GP) \cite{koza}. In order to construct the model that contains similar knowledge as that used by the stratigrapher to classify 5 different deposition types, Step 5 calculates the attribute values listed in Table \ref{attributes} for every depositional unit labeled by the stratigrapher.

\begin{table}
\caption{Attributes calculated for each depositional unit.}
\begin{center}
\begin{tabular}{|c|c|c|}
\hline
\multicolumn{1}{|c|}{symbol\%}
& \multicolumn{1}{|c|}{symbol thickness}
& \multicolumn{1}{|c|}{symbol max}
\\
\hline
\hline
 a\% & a\_ thickness &  a\_max  \\ \hline
 ab\% & ab\_thickness &  ab\_max  \\ \hline
 ba\% & ba\_thickness &  ba\_max  \\ \hline
 b\% & b\_thickness &  b\_max    \\ \hline
 bc\% & bc\_thickness & bc\_max    \\ \hline
cb\% & cb\_thickness &  cb\_max    \\ \hline
 c\% & c\_thickness &  c\_max    \\ \hline
 cd\% & cd\_thickness &  cd\_max      \\ \hline
 dc\% & dc\_thickness &  dc\_max     \\ \hline
 d\% & d\_thickness&  d\_max  \\ \hline
 variation &  total\_thickness &  no\_segments \\ \hline
\end{tabular}
\vspace{-0.4cm}
\label{attributes}
\end{center}
\end{table}

A depositional unit can contain 1 or more segments, hence is represented by 1 or more Vsh symbols. For example, an unit labeled as deposition type \texttt{A} can contain 4 symbols \texttt{a,ab,b,a}. 
We used the thickness of each symbol to calculate the attribute values in Table \ref{attributes}.  There, the "symbol\%" column gives the percentage of each symbol's thickness over the total thickness of the unit. For example, ``a\%`` is the percentage of the thickness of  ``a`` symbols over the total thickness of the unit. Similarly, the ``symbol thickness`` column gives the accumulated thickness of each symbol in the unit. The "symbol max" column gives the maximum thickness of each symbol in the unit. 

Two attributes that are used to estimate the smoothness of the GR readings in the unit are ``no\_segments``, which is the number of segments in the unit) and "variation", which is the average \textit{distance} of the neighboring symbols in the unit. Here, \textit{distance} is defined as the number of jumps between 2 symbols. For example, the distance between symbols $a$ and $dc$ is 8. Variation of symbol sequence $a$, $ba$, $dc$ is (2+6)/2=4. Smoothness of GR readings is an important indication of the deposition type of the unit (see Table \ref{framework}).

Step 6 uses these attribute information and their associated depositional labels to train 5 classifiers (see Section \ref{co-evolution} for implementation details).
In Step 7, the 5 classifiers, in addition to the Vsh symbols and their associated thickness, are then used to train a FST as the final model (see Section \ref{fuzzyfst} for the training process details). 
The final FST model takes a sequence of Vsh symbols representing GR readings and their associated thickness as inputs.
It produces a sequence of depositional labels based on the decisions made by the 5 classifiers.  

To apply the trained FST model to interpret new GR log data, the GR log data are first transformed into fuzzy symbols following Step 1, 2 and 4. 
Acting like a stratigrapher, the FST model interprets these symbols and assigns their associated depositional labels.

\section{Data Set}\label{case}
One GR well log data set from a deepwater reservoir was provided for us to develop and test our methodology.
The GR readings consist of 6,150 data points, each of which was taken at a half foot interval between 4,200 and 7,200 feet under the earth's surface. A stratigrapher has examined the data and assigned 50 depositional labels at different depth level. 
Following the workflow in Figure \ref{flow}, we first converted the data into Vsh values.
The segmentation of the Vsh data is explained in the following section.

\section{Segmentation of Vsh Data}\label{segmentation}
The segmentation process consists of two steps (Step 2 and 3 in the workflow). The first step involves partitioning Vsh data into segments and using the mean value of the data points within the segment to represent the original data. In order to accurately represent the original data, each segment is allowed to have arbitrary length. In this way, areas where data points have low variation will be represented by a single segment while areas where data points have high variation will have many segments.

There are various approaches to segment data. Due to budget constraint, we only adopted one \cite{lin_keogh_lonardi_chiu} and modified it to suit our project.  This segmentation method starts by having one data point in each segment. That is the number of segments is the same as the number of original data points. Step-by-step, neighboring segments (data points) are gradually combined to reduce the number of segments. This process stops when the number of segments reaches the predetermined number of segments. 

At each step, the segments whose merging will lead to the least increase in $error$ are combined. The $error$ of a segment is defined as:

\begin{equation}
error =\displaystyle\sum_{i = 1}^{n} (d_i - \mu)^2
\label{eq:eq2}
\end{equation}
where $n$ is the number of data points in the segment, $\mu$ is the mean value of the segment, $d_i$ is the $i$th data value in the segment. 

\begin{figure}
\begin{minipage}[t]{0.49\linewidth}
\centerline{\includegraphics[width=4.5cm,height=3.5cm]{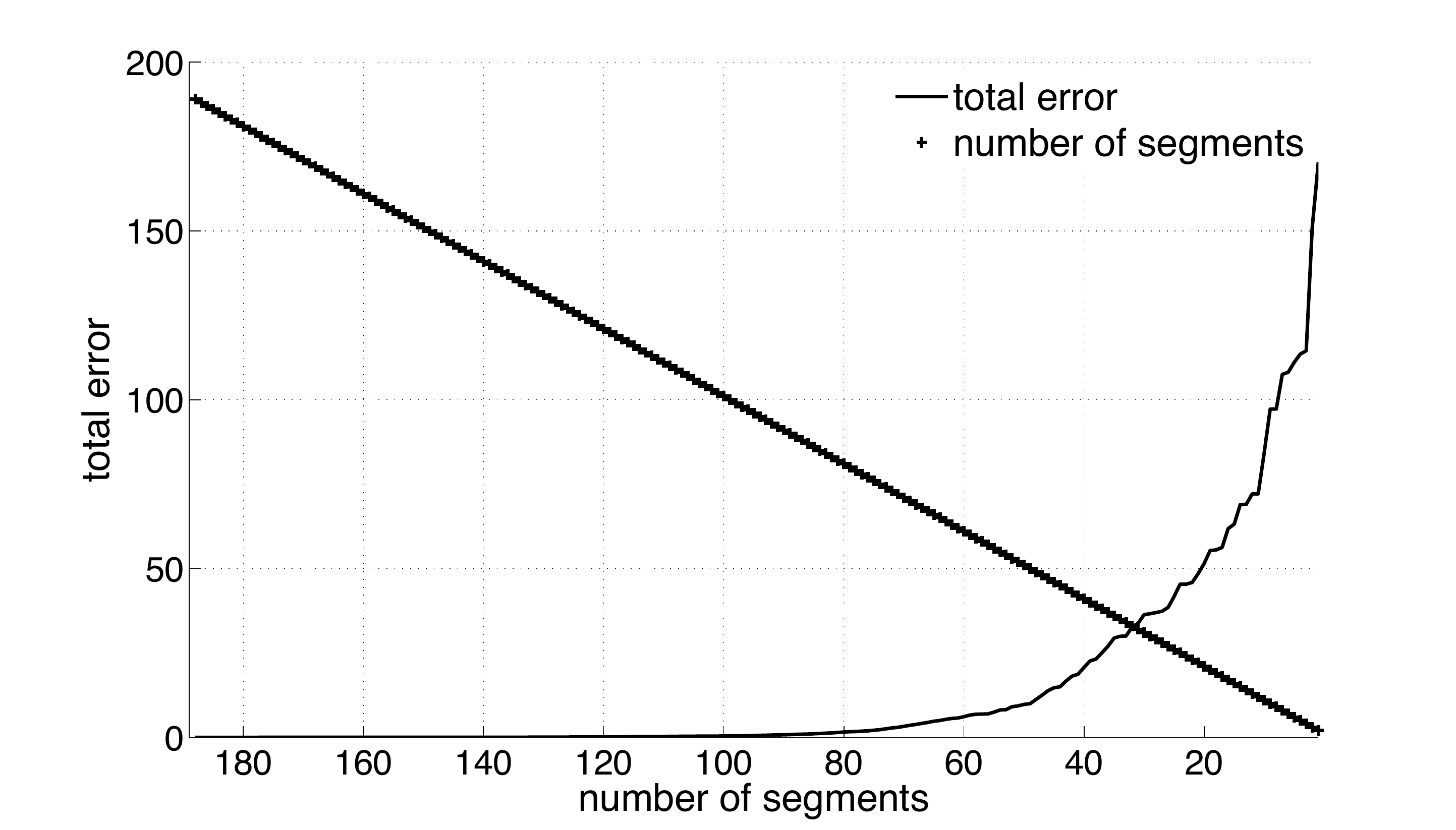}}
\end{minipage}
\begin{minipage}[t]{0.49\linewidth}
\centerline{\includegraphics[width=4.5cm,height=3.5cm]{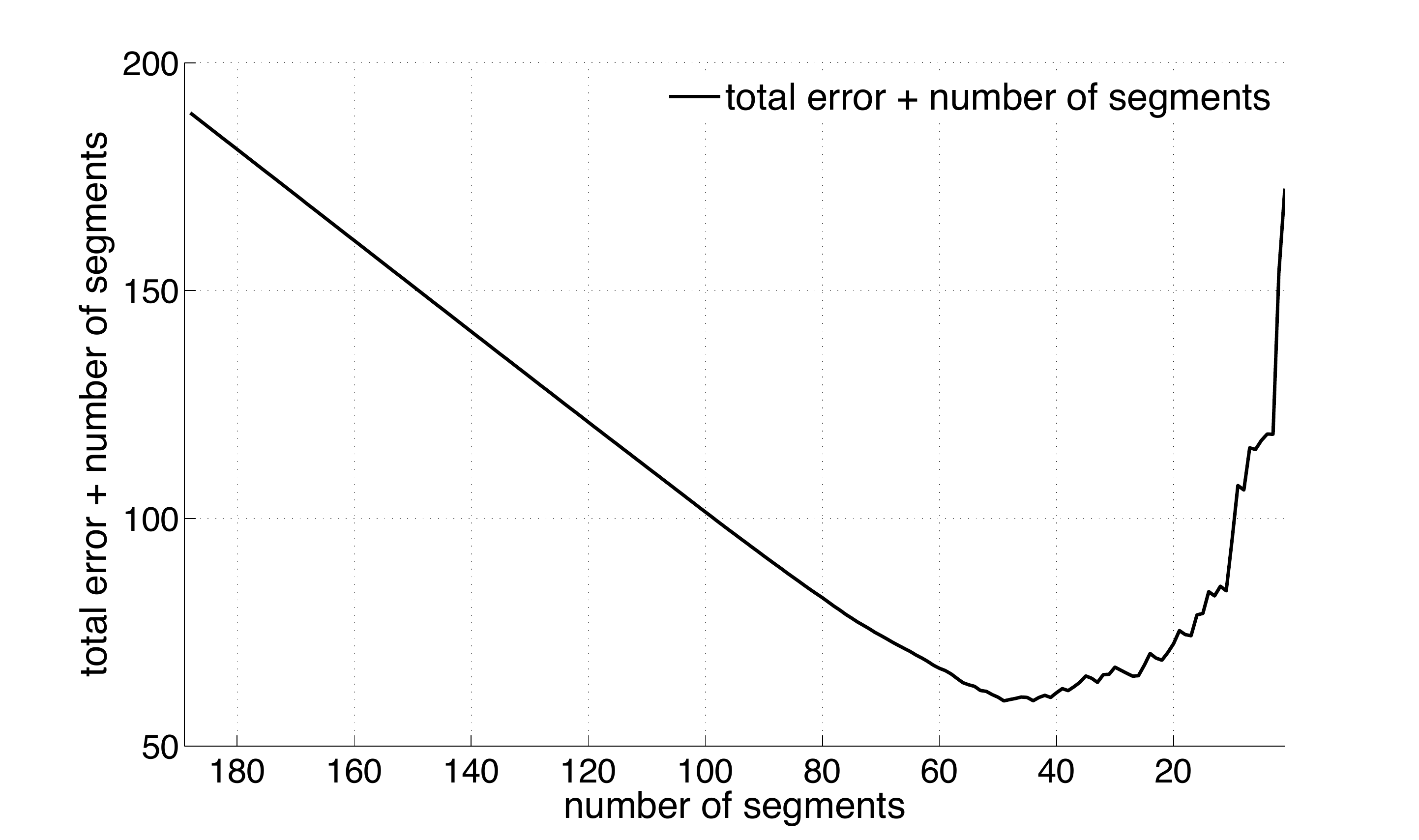}}
\end{minipage}
\caption{Number of segments vs. total error (left); A compromised solution (right).}
\label{conflict}
\end{figure}

Although a larger number of segments capture the data trend better, it is also more difficult to interpret. Ideally, we want to use the smallest possible number of segments to capture the trend of the log data. 
Unfortunately, these two objectives are in conflict: the total $error$ of all segments monotonically increases as the number of segments decreases (see Figure \ref{conflict} (left)). We therefore devised a compromised solution where a penalty is paid for increasing the number of segments. Equation \ref{eq:eq3} gives the new $error$ criterion which is defined as the total error defined previously $plus$ the number of segments, $N$.

\begin{equation}
f = N + \displaystyle\sum_{i = 1}^{N}error_i 
\label{eq:eq3}
\end{equation}

During the segmentation process, the above $f$ function is evaluated at each step. As long as $f$ value is decreasing, the system continues to merge segments. Once $f$ starts increasing, it indicates that farther reducing of the number of segments will sacrifice log characteristics, hence the segmentation process terminates. Figure \ref{conflict} (right) shows 50 is the optimal number of segments under this compromised approach.

Although simple, the proposed compromised approach to decide the number of segments may sacrifice the the capacity to capture
the data trend. We will discuss possible improvement and future work in Section \ref{discussions}.

Based on the described scheme, a computer system was developed to carry out the segmentation task. 
Computer generated segments, however, are not always aligned with the data points where stratigrapher assigns depositional labels. 
The second step of the segmentation process (Step 3) adjusts the boundary locations of the depositional units to fix the edge effect. 
In most cases, this requires shifting the labels up or down a few data points.

\section{Transform Vsh Data to Fuzzy Symbols}\label{fuzzy}
Segmented Vsh data are represented as a set of numerical values, $\overline{Vsh}={\overline{s_1},\overline{s_2},\overline{s_3}\dots}$, where $\overline{s_i}$  is the mean value of the data within the $i$th segment. Step 4 assigns each segment with one of the symbols {$a$, $b$, $c$, $d$} according to the following rule:

\begin{equation}
symbol_i= \left\{ \begin{array}{ll} a, & \overline{s_i} < 0.3;\\
b, & 0.3\le \overline{s_i} < 0.5;\\
c, & 0.5\le \overline{s_i}\le 0.7;\\
d, & \overline{s_i} > 0.7.
\end{array}\right.
\label{eq:eq4}
\end{equation}

The cut points (0.3, 0.5 and 0.7) are provided by the stratigrapher who interpreted the GR log. 

While some segments are clearly within the boundary of  a particular symbol region, others may not be so clear. In this case, a crisp symbol does not represent its true value. By contrast, fuzzy symbols use membership function to express the segment can be interpreted as both symbols with some possibilities. Fuzzy symbols are therefore more expressive in this case.

In fuzzy logic \cite{zadeh}, a membership function (MF) defines how each data point in the input space is mapped into a membership value (or degree of membership) between 0 and 1. The input space consists of all possible input values. In our case, Vsh data have values between 0 and 1. Since we adopted 4 symbols to represent Vsh data, 4  trapezoidal-shaped MFs were designed, one for each of the 4 symbols.  

We chose trapezoidal-shaped MF based on our previous experiences \cite{yu_wilkinson1} and \cite{yu_wilkinson2}, where we found trapezoidal-shaped MF is suitable for
the transformation of various types of well log, such as porosity, density and sonic log. 

\begin{figure}
\centering
\includegraphics[height=1.2in, width=3in]{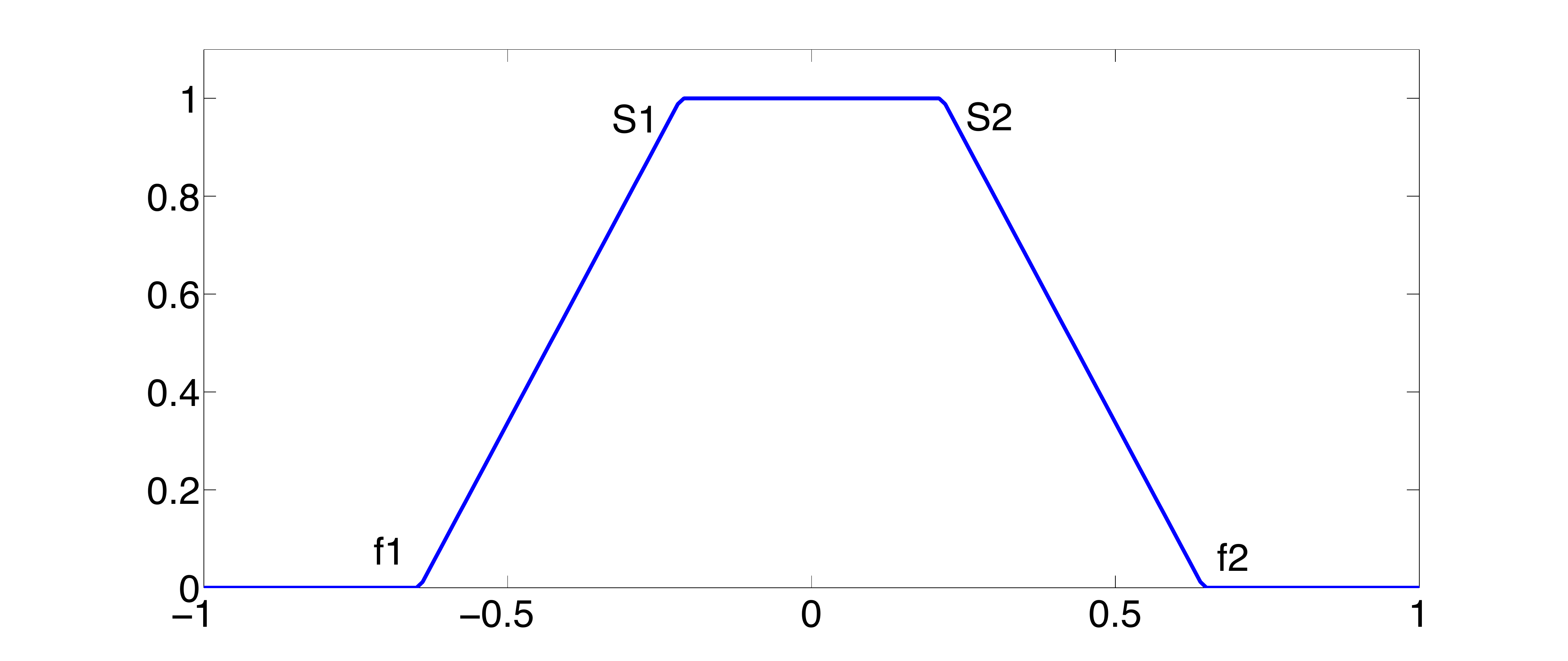}
\caption{f1, f2, s1, s2, define a trapezoidal-shaped MF.}
\label{trapmf}
\end{figure}

To design a trapezoidal-shaped MF, 4 parameters are required: $f_1$ and $f_2$ are used to locate the `feet` of the trapezoid and $s_1$ and $s_2$ are used to locate the `shoulders` (see Figure \ref{trapmf}). 
We designed these four parameters in the following way. Let $t_1$ and $t_2$ be the two cut points that define symbol $n$ and $t_2 > t_1$: 
\begin{center} 
$y= \displaystyle\frac{t_2-t_1}{4}$\\
$f_1=t_1-y; s_1=t_1+y; s_2=t_2- y; f_2=t_2+y$\\
\end{center} 

There are two exceptions: symbol $a$ has $f_1$ = $t_1$ and symbol $d$ has $f_2$ = $t_2$.
Table \ref{mf} gives the four parameters for the 4 MFs.

\begin{table}
\centering
\caption{The 4 trapezoidal-shaped MF parameter values.} 
\begin{tabular}{|c|c|c|c|c|} \hline
symbol & $f_1$ & $s_1$ & $s_2$ & $f_2$ \\ \hline
 a &0&	0.075&0.225&0.375 \\ \hline
 b &0.25&0.35&0.45&0.55\\ \hline
 c & 0.45&0.55&0.65&0.75  \\ \hline
 d &0.625	&0.775&	0.925&	1  \\ 
\hline\end{tabular}
\label{mf}
\end{table}
Once the 4 parameters are decided, the MF $f$ for input $x$ is defined in Equation \ref{4_para}.
With that, 4  trapezoidal shaped  MFs ($MF_a$, $MF_b$, $MF_c$, $MF_d$)  were defined to transform Vsh data into fuzzy symbols. \\
\begin{equation}
f(x,f_1,f_2,s_1,s_2)= 
\begin{cases} 
0, & \mbox{if $x \le f_1$} \\
\frac {x-f_1}{s_1-f_1}, & \mbox{if  $f_1\le x \le s_1$} \\
1, & \mbox{if $s_1 \le x \le s_2$}\\
\frac {f_2-x}{f_2-s_2}, & \mbox{if $s_2\le x \le f_2$} \\
0, & \mbox{if  $f_2 \le x$}
\end{cases} 
\label{4_para}
\end{equation}

To fully realize the power of fuzzy logic in enhancing the FST interpretation ability of noisy and uncertain GR data, both the classification rules in Step 6 and the FST in Step 7 have to manipulate fuzzy symbols. Previously, we have implemented fuzzy rules to classify other type of well log data [Yu and Wilkinson, 2007].   
However, in this first version of the interpretation system, we started with a crisp version of classification rules and FST model.
This is because the fuzzy version of the interpretation system is too complicated for a stratigrapher to comprehend.
A simpler crisp version is an ideal first step to introduce the computer interpretation system to stratigraphers and win their acceptance.
We will discuss implementing the fuzzy interpretation system in Section \ref{discussions}.

We used the designed 4 MFs to map each segment to one of the following 10 crisp symbols ($a$, $ab$, $ba$, $b$, $bc$, $cb$, $c$, $cd$, $dc$, $d$) in the following ways. 
If a segment $\overline{s_i}$ belongs to the region symbol $j$ 100\%, i.e. $f(\overline{s_i} ,MF_j) = 1$, symbol $j$ is used to represent that segment.
If a segment belongs to two symbol regions $j$ and $k$, we evaluated its degree of membership to the 2 symbol regions.
If $f(\overline{s_i} ,MF_j) > f(\overline{s_i}, MF_k)$, the segment is mapped to symbol $jk$. Otherwise, it is mapped to $kj$.

We implemented the segmentation method described in Section \ref{segmentation} and the symbols transformation algorithm explained in this section in Matlab and applied the software to the Vsh data. The results are presented in the following subsection.

\subsection{Results}
\begin{figure}[htp]
\centerline{\includegraphics[width=4in,height=2.3in]{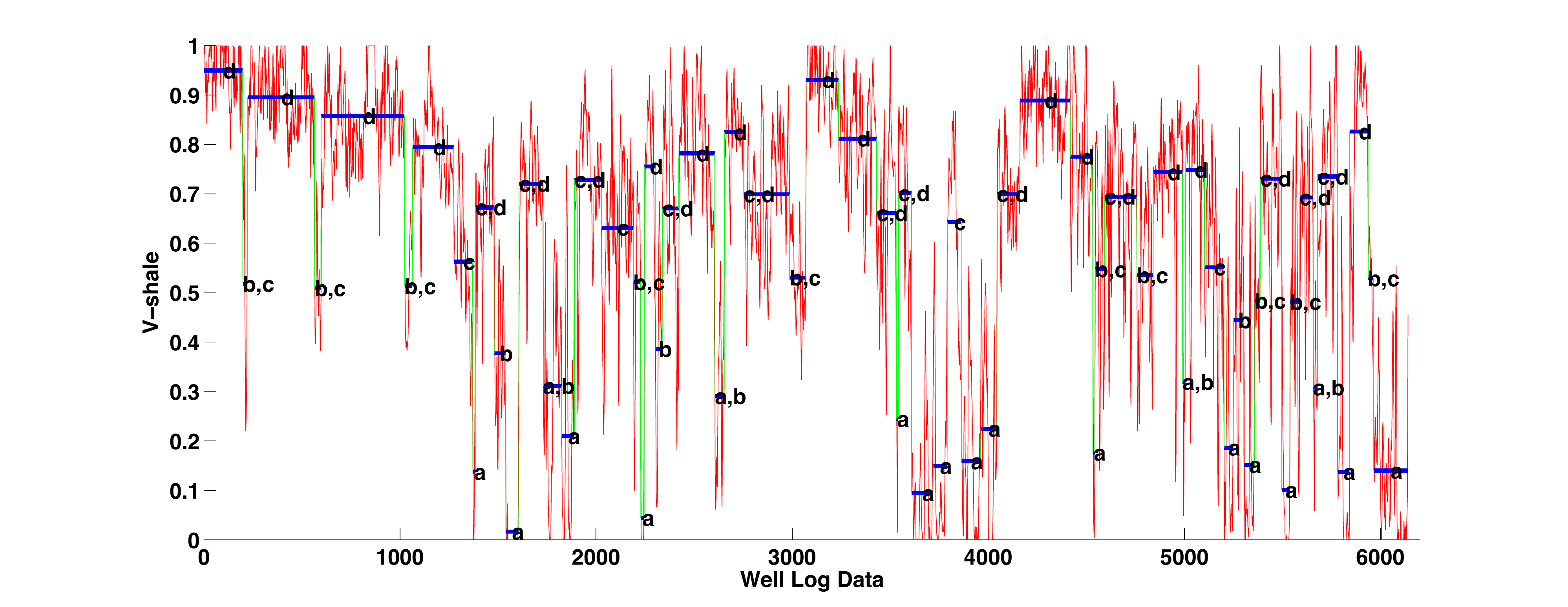}}
\caption{Vsh data transformed into 62 symbols.} 
\label{fig2}
\end{figure}

The 6,150 Vsh data points were segmented and mapped into 62 symbols. They are shown in Figure \ref{fig2}.

We aligned the decision points of the 62 symbols with the positions where the stratigrapher assigned the 50 depositional labels. In the case where one symbol region (segment) has more than one depositional labels assigned to it, we divided the region into multiple segments according to where the position of the label. After this process, the total number of symbol regions (segments) increased from 62 to 82.

The number of symbols (segments) in each depositional unit varies, ranging from 1 to 5.
Also, some symbols represent a larger number of data points (thicker) than others. 
This information is important in determining their deposition types. 
Step 5 calculates the attribute values listed in Table \ref{attributes} for each of the 50 depositional units. 
The resulting data are then used to train 5 classifiers described in the following section. 

\section{Co-Evolution of Classifiers}\label{co-evolution}
Step 6 trains 5 classification rules to identify 5 different deposition types.
Among the 50 depositional units, 4 are A, 9 are OA, 14 are M, 19 are OB and 4 are MTC. If we train each classifier individually as a binary classifier, it would be very difficult to obtain good results since the number of depositional units for one class is too small to balance against the number of depositional units for the 4 other classes. An alternative approach is to train the 5 binary classifiers simultaneously. The co-evolutionary computation system described in \cite{yu_wilkinson2} was developed based on this concept.

\begin{figure}[htp]
\vspace{-0.5cm}
\centerline{\includegraphics[width=3.6in, height=2.5in]{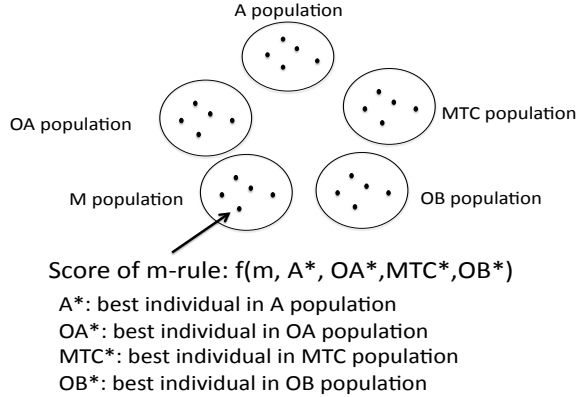}}
\vspace{-0.5cm}
\caption{The co-operative co-evolution model.}
\label{two_pops}
\end{figure}

In this co-evolutionary system, each binary classifier is represented as a Boolean rule tree.  Five populations, each trains one of the five classifiers,  are evolved independently and simultaneously. To encourage the co-operation of the 5 populations to evolve the best \emph{team}, the fitness of an evolved individual rule in one population is determined by how well it collaborates with the $best$ rules evolved in the other 4 populations to perform the overall classification task. Here, the $best$ rule is the one with the best classification accuracy in the population, based on the fitness function defined in Section \ref{distance}. This co-operative co-evolution model is based on \cite{potter_dejong} and is illustrated in Figure \ref{two_pops}.

In terms of implementation, a rule from one population is combined with the best rules from the 4 other populations using an if-then-else template, such as the following: \\

\texttt{if (OA-rule is evaluated as True)\\ 
\indent \indent then OA \\
\indent else if (A-rule is evaluated as True)\\  
\indent \indent then A \\
\indent else if (MTC-rule is evaluated as True)\\  
\indent \indent then MTC \\
\indent else if (OB-rule is evaluated as True)\\  
\indent \indent then OB \\
\indent else M.}\\
The performance of this combined team determines the fitness of the rule in the population.

At generation 0 when no best rule is known, a rule is randomly selected from each population as the best rule for that population. After that, the best rule is updated at the end of every generation based on the fitness evaluation, so that a good rule can be immediately used to combine with rules in other populations in the following generation and impact evolution. 

With 5 rules in each team, there are 5! ways to combine (order) them using the if-then-else template. Since rules applied earlier in the if-then-else template are weighted more than rules that are applied later in the classification decision, it is important to acquire a suitable rule sequence for a team to achieve good classification accuracy.

To learn the best rule order strategy, we used a simple hill-climbing approach. At the beginning of each experimental run, a random order of rule sequence is generated (e.g. OA, A, MTC, OB, M). This order is used to combine rules for fitness evaluation at generation 0. At the end of generation 0, the best rule from each of the 5 populations are selected. Meanwhile, the current order sequence is mutated to obtain a new rule order sequence. If the team of the 5 best rules combined using the new order sequence gives better or equal classification accuracy than that produced by the team combined based on the old sequence, the new sequence is adopted and used in the next generation for rule combination. Otherwise, the old sequence is retained and reused in the next generation. This process is repeated at the end of each generation to determine the rule order strategy in the next generation.

\subsection{Experimental Setup}\label{distance}
We implemented the co-evolutionary system using a GP software called PolyGP \cite{yu_clack}. GP \cite{koza} applies genetic algorithms \cite{holland} to evolve program trees. In our case, the program trees are Boolean classification rules. When a Boolean rule is executed, it returns a value of $true$ or $false$. If the value is $true$, the classification is positive. Otherwise, the classification is negative. 

We provided six operators for GP to construct the classification rules: $and$, $or$, $nor$, $nand$, $<$ and $>$. They can combine any subset of the  33 attributes listed in Table \ref{attributes} and the following constants to form a Boolean rule:
\begin{itemize}
\item Boolean constants: \texttt{true} and \texttt{false};
\item integer constants: 1 -- 10;
\item percentage constants:  0.0 -- 1.0;
\item double constants: 0.0 -- 250.0;
\end{itemize}

Figure \ref{a-rule} gives an example of a Boolean rule that classifies deposition type $A$.

The attributes and the constants have different types, while each operator can only be applied to a certain type (see Table \ref{type}).

\begin{table}[h]
\caption{Operators, attributes, constants and their types.}
\begin{center}
\begin{tabular}{|c|c|}
\hline
\multicolumn{1}{|c|}{name}
& \multicolumn{1}{|c|}{type}
\\
\hline
 \hline
and &$bool \rightarrow bool \rightarrow bool$  \\ \hline
or &$bool \rightarrow bool \rightarrow bool$  \\ \hline
nand &$bool \rightarrow bool \rightarrow bool$  \\ \hline
nor &$bool \rightarrow bool \rightarrow bool$  \\ \hline
$<$ & $a \rightarrow a \rightarrow bool$ \\ \hline
$>$ & $a \rightarrow a \rightarrow bool$ \\ \hline
symbol\%& $percentage$\\ \hline
symbol\_thickness& $double$ \\ \hline
symbol\_max & $double$ \\ \hline
variation & $integer $  \\ \hline
no\_segments  & $integer$ \\ \hline
\% constants & $percentage$\\ \hline
integer constants & $integer$\\ \hline
double constants & $double$ \\ \hline
boolean constants & $bool$ \\ \hline
 \end{tabular}
\label{type}
\end{center}
\end{table}

To assure only semantically meaningful rules are generated, the PolyGP employed a type system to performs type checking.
For example, $<$ can be applied to compare attributes with the same type, e.g. \texttt{d\%} with \texttt{b\%} , \texttt{a\_thickness} with \texttt{d\_thickness} and so on. 
The $a$ in Table \ref{type} is a type variable which can be substituted with any concrete type, such as \texttt{percentage} and \texttt{integer}. 

\begin{figure}[htp]
\vspace{-1.0cm}
\centerline{\includegraphics[width=3.8in, height=2.8in]{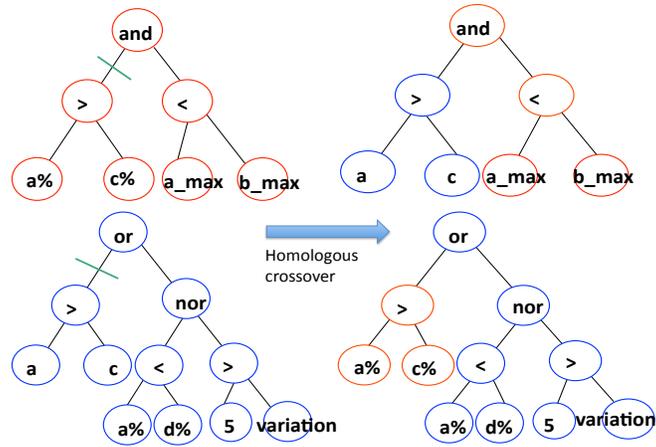}}
\vspace{-0.8cm}
\caption{The homologous crossover operator.}
\label{crossover}
\end{figure}

To work with the Boolean rule tree representation, we employed four genetic operators in this study: homologous crossover, $and$-crossover, $or$-crossover and mutation. Homologous crossover selects common location in both parent trees to carry out the crossover operation and produce two offspring (see Figure \ref{crossover}). 

\begin{figure}[htp]
\vspace{-0.8cm}
\centerline{\includegraphics[width=3.8in, height=2.8in]{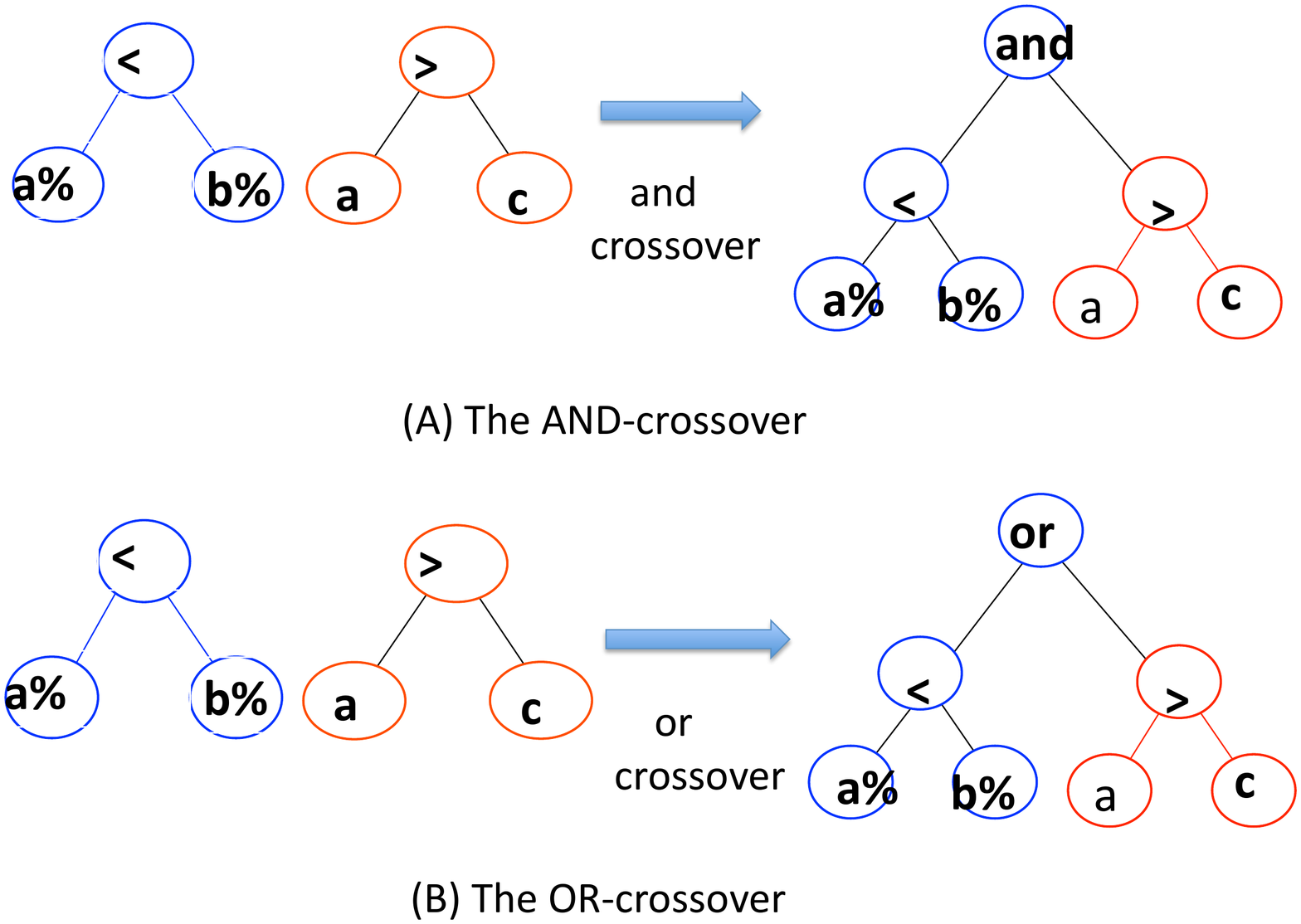}}
\vspace{-0.8cm}
\caption{(A)$and$-crossover (B) $or$-crossover. }
\label{orcrossover}
\end{figure}

\begin{figure}[htp]
\vspace{-1cm}
\centerline{\includegraphics[width=3.8in, height=2.8in]{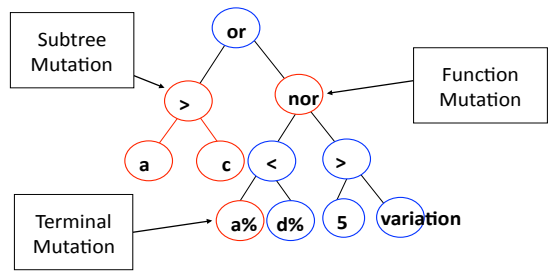}}
\vspace{-2.1cm}
\caption{The mutation operator. }
\label{mutation}
\end{figure}

The $and$-crossover combines two parent rules into one rule using the $and$ operator (see Figure \ref{orcrossover}(A)). The $or$-crossover combines two parent rules into one rule using the $or$ operator (see Figure \ref{orcrossover}(B)). The mutation operation can perform sub-tree, function and terminal mutations, depending on the selected mutation location (see Figure \ref{mutation}). 
Mutation produces single offspring.

Table \ref{para} lists the GP parameter values used to conduct the experimental runs. 
\begin{table}[h]
\caption{GP parameters for experimental runs.}
\begin{center}
\scalebox{0.85}{
\begin{tabular}{|c|c|c|c|}
\hline
\multicolumn{1}{|c|}{parameter}
& \multicolumn{1}{|c|}{value}
& \multicolumn{1}{|c|}{parameter}
& \multicolumn{1}{|c|}{value} 
\\
\hline
 \hline
tournament selection & size 2 &  homoXover rate & 30\%\\ \hline
max tree depth& 6 &  orXover rate & 30\%   \\ \hline
population size  & 100&  mutation rate &  30\% \\ \hline
max generation & 200 &  andXover rate & 30\% \\ \hline
number of runs & 50 & elitism & 1 \\ \hline
\end{tabular}}
\label{para}
\end{center}
\end{table}
A population consists of 100 Boolean rules.
At the beginning of each new generation, the rule with the best fitness in the current population is copied over to the new population (elitism = 1).
The rest 99 Boolean rules are then generated by selection and genetic operations as described below.

A rule is first selected (using tournament selection of size 2) from the current population. 
Next, one of the 4 genetic operations is applied based on the following probabilities:

\vspace{0.2cm}
\texttt{if (random() < 30\%)\\
\indent \indent perform homologous crossover\\
\indent else if (random() < 30\%)\\
\indent \indent perform mutation\\
\indent else if (random() < 30\%)\\
\indent \indent perform orCrossover\\
\indent else if (random() < 30\%)\\
\indent \indent perform andCrossover\\
\indent else\\
\indent \indent perform copy}
\vspace{0.2cm}

To carry out any of the 3 crossover operations, another rule is selected using the same tournament method.

The fitness of a rule is the classification accuracy of the combined if-then-else rule team that the rule is a part of. 
When the team gives a correct interpretation for a depositional unit, it is a hit. 
The number of hits divided by the number of depositional units in the training data (50) is the fitness of the evaluated rule. 

Additionally, we used penalty to discourage generating large size rules.
Initially, all rules at generation 0 have the same depth of 6, hence $2^6-1= 63$ nodes. After that, the rule trees tend to grow in size.
A rule whose tree size ($l$) is larger than 150 is penalized.
The final fitness of a rule is therefore:

\begin{center}
\begin{tabular*}{0.4\textwidth}{@{\extracolsep{\fill}}l p{6cm}}
$fitness =  \displaystyle \frac{\sum_{i = 1}^{N} hit_i}{N}-(\frac{l-150}{l})^2,$ & N=50
\end{tabular*}
\end{center}

Among the 5 depositional types, A and OA are close to each other while M, OB and MTC are similar in their geological characteristics. To promote rules that produce less serious misclassification, the distance between the target label and the classified label is used as the secondary criterion for selection. Here, the 5 depositional labels are ordered as follows: A, OA, M, OB, MTC. The \textit{distance} between two depositional labels is the number of jumps between them. For example, the distance between A and OA is 1 and the distance between A and MTC is 4. If two rules have the same fitness value, the one that gives a smaller distance error is the winner during tournament selection.  



\subsection{Results}
\begin{figure}
\centerline{\includegraphics[width=3.4in]{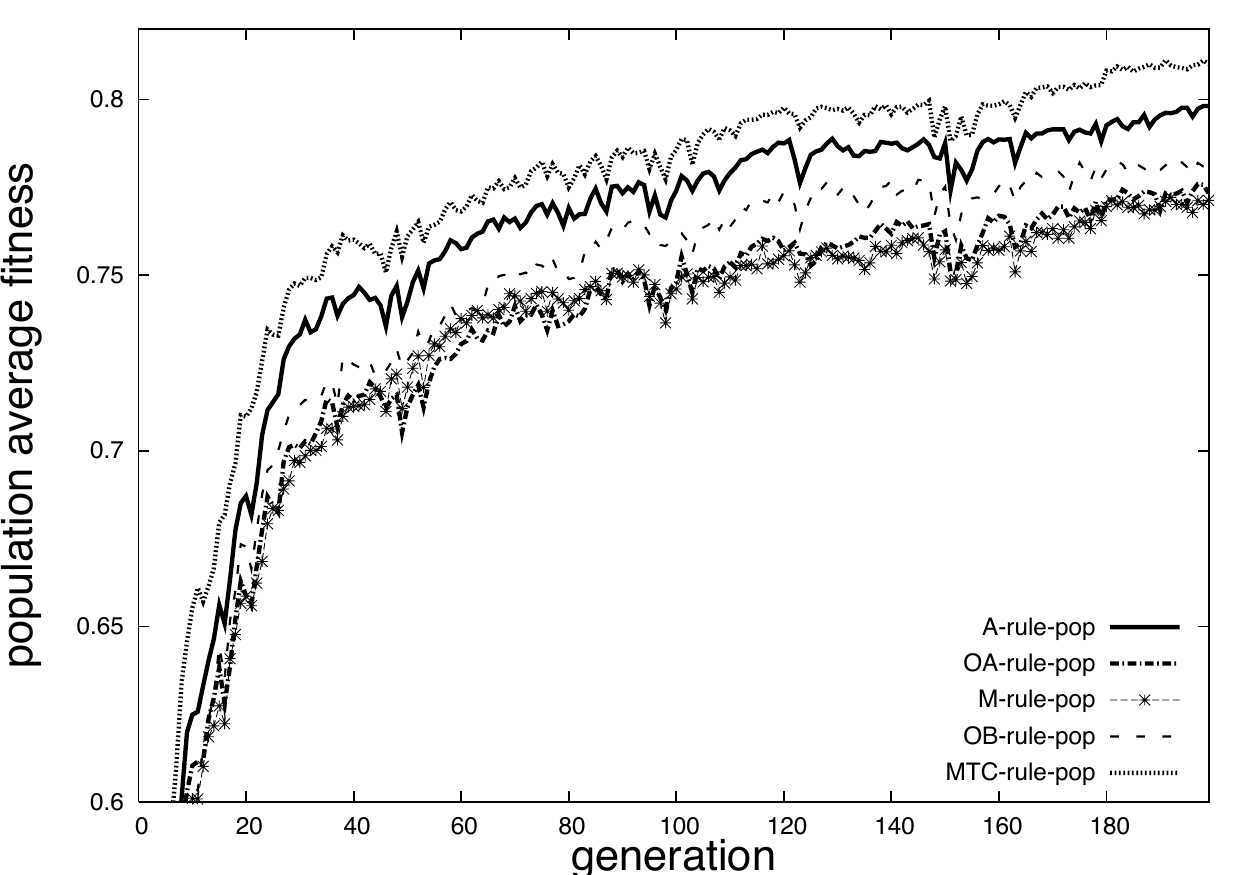}}
\caption{Average fitness of the 5 co-evolved populations.} 
\label{avg}
\end{figure}

We performed 50 experimental runs. Figure \ref{avg} gives the average fitness of the 5 co-evolved populations. It shows that all 5 populations improved consistently throughout the runs. Among them, MTC-rule population has the highest average fitness. This is followed by the A-rule and OB-rule populations. The M-rule and OA-rule populations have the lowest average population fitness.

\begin{figure}
\centerline{\includegraphics[width=3.4in]{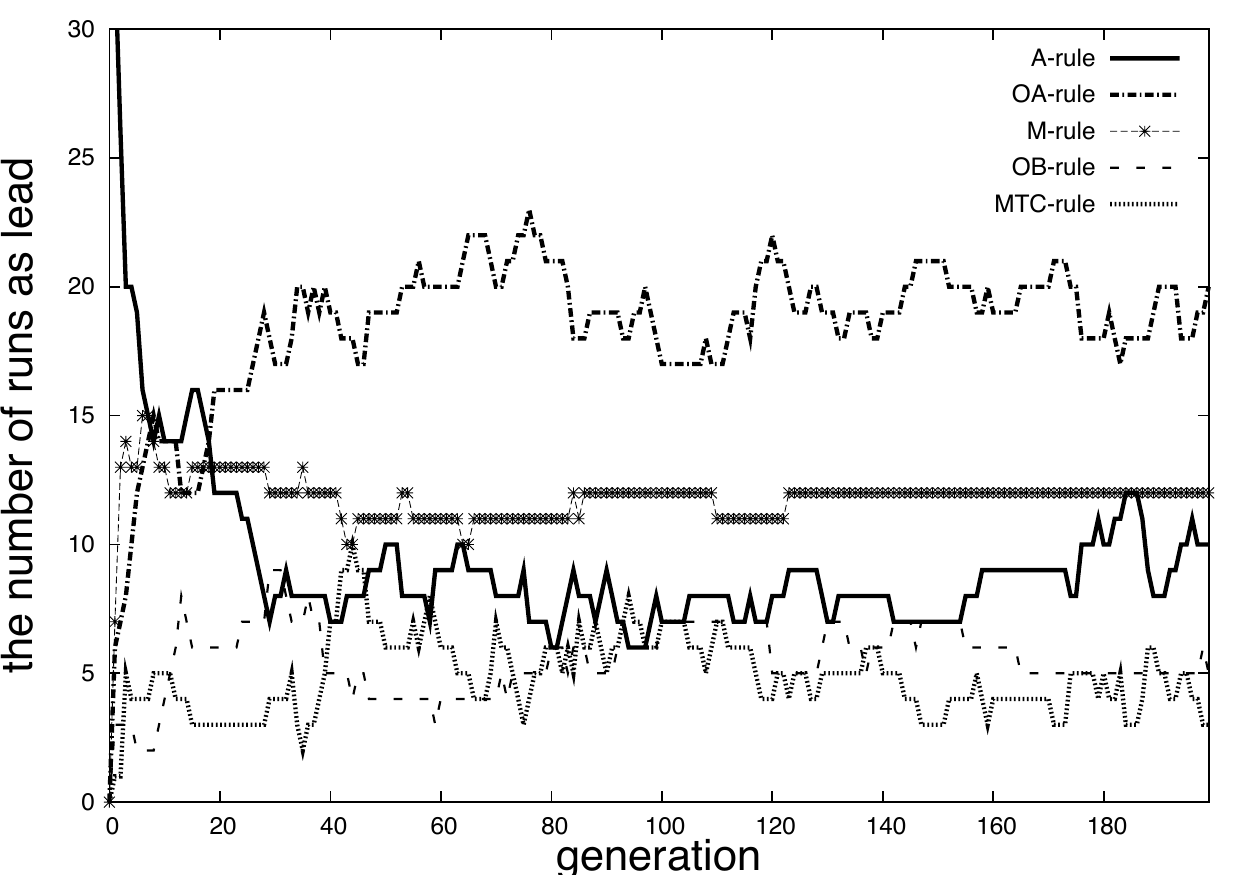}}
\caption{The number of runs that each rule is selected as the lead.} 
\label{lead}
\end{figure}

These results suggest that MTC-rule has the least impact on the overall classification results. Once $best$ rules from other populations are integrated into a team, any MTC-rule in the population can produce good classification results.  We examined the number of runs that each rule was selected as the lead classifier in the team (see Figure \ref{lead}) and found that MTC-rule is the least selected one, indicating that it is of the least impact. This result is consistent with our hypothesis. 

The two most selected rules as the lead classifier in the team are OA-rule and M-rule. Figure \ref{avg} also shows that these two populations have the lowest average fitness. This indicates that the $best$ rules from other populations are not sufficient for the team to produce good classification accuracy; it is critical for a team to have good OA and M rules to produce good classification results. In other words, OA-rule and M-rule have more impact  on the overall classification results than other rules. 

One factor that might have contributed to this evolutionary dynamics is the number of training data in each class.
In general, training process biases toward the class that has a larger number of training data. 
As a result, that classification rule has a higher impact on the overall classification results. 
In this data set, MTC class has the smallest number (4) of data points while OA and M classes have the largest number (14 and 19).
It is understandable that the training process ignored MTC-rule but focusing on evolving good OA and M rules for the team to produce good overall classification accuracy.

The best team produced from the 50 runs gives classification accuracy of 90\%: 5 of the 50 depositional units were mis-classified.
Table \ref{matrix} gives the details. 

\begin{table}[h]
\caption{Classification accuracy of the best team.}
\scalebox{1.0}{
\begin{tabular}{|c|c|c|c|c|c||c|c|c|c|c|}
\hline
\multicolumn{1}{|c|}{}
& \multicolumn{1}{|c|}{A}
& \multicolumn{1}{|c|}{OA}
& \multicolumn{1}{|c|}{M}
& \multicolumn{1}{|c|}{OB} 
& \multicolumn{1}{|c|}{MTC} 
& \multicolumn{1}{|c|}{A} 
& \multicolumn{1}{|c|}{OA} 
& \multicolumn{1}{|c|}{M} 
& \multicolumn{1}{|c|}{OB} 
& \multicolumn{1}{|c|}{MTC} 
\\
\hline
\hline
A&4&0 &0&0&0&100\%&0\%&0\%&0\%&0\%\\ \hline
OA& 0 & 9& 0&0&0&0\%&100\%&0\%&0\%&0\%   \\ \hline
M &0&0&13&1&0&0\%&0\%&93\%&7\% &0\% \\ \hline
OB & 0 &  0 & 1&18&0&0\% &0\%&5\%&95\%&0\%\\ \hline
MTC & 0& 0&1&2&1&0\%&0\%&25\%&50\%&25\% \\ \hline
\end{tabular}
}
\label{matrix}
\end{table}
\normalsize
Among the 5 mis-classified depositional labels, two MTC were mis-labeled as OB, one MTC was mis-labeled as M, one OB was mis-labeled as M and one M was mis-labeled as OB. As mentioned previously, M, OB and MTC have similar depositional ingredients. These mis-classifications, hence, are not considered to be serious.

\begin{figure}[htp]
\vspace{-0.4cm}
\centerline{\includegraphics[width=3.6in]{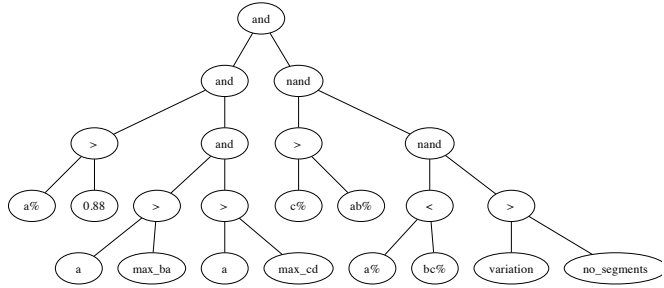}}
\vspace{-2.5cm}
\caption{The ``A`` classification rule.} 
\label{a-rule}
\end{figure}

\begin{figure}[htp]
\centerline{\includegraphics[width=3.8in]{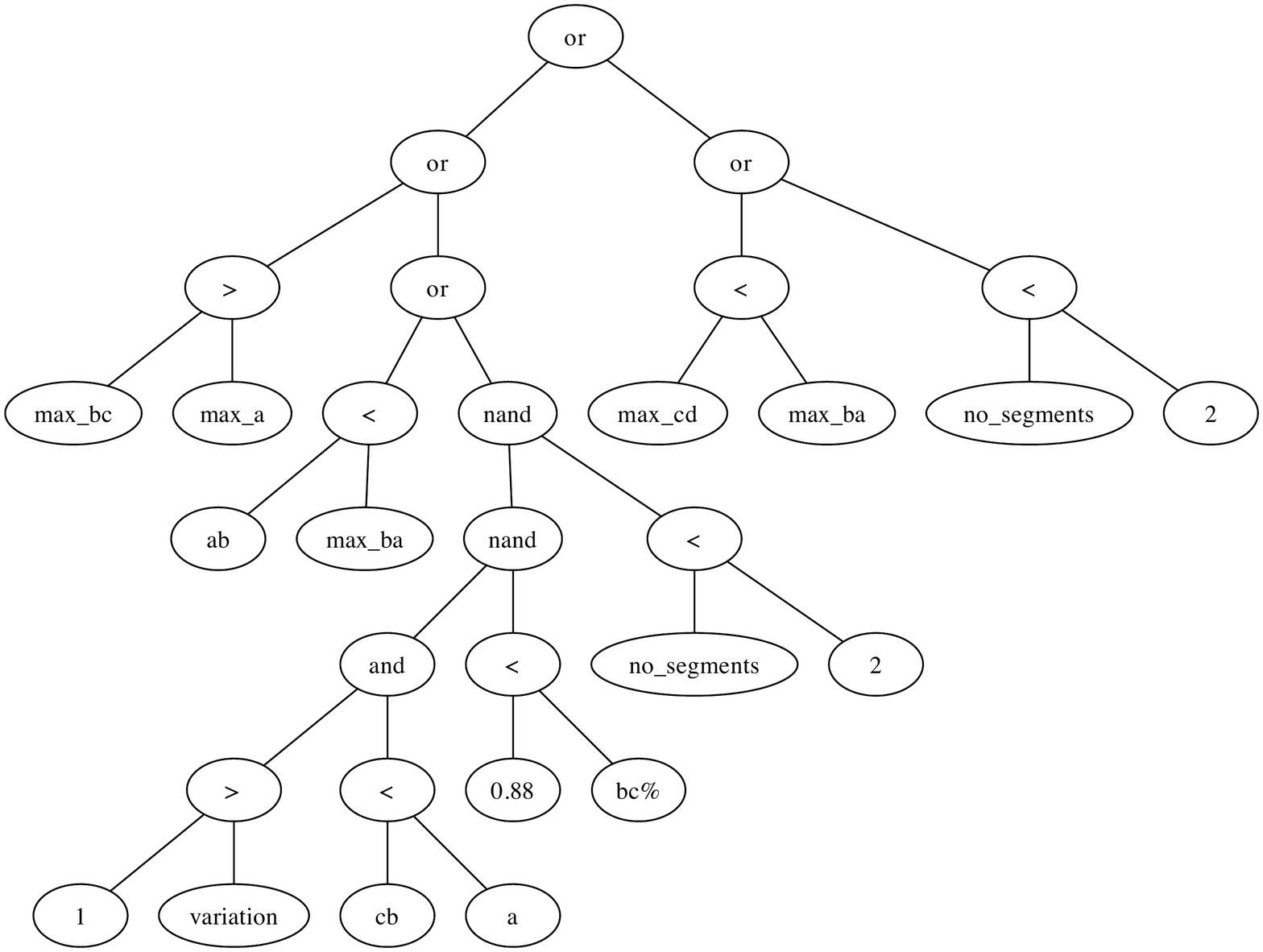}}
\caption{The ``OA`` classification rule.} 
\label{oa-rule}
\end{figure}

\begin{figure}[htp]
\vspace{-0.4cm}
\centerline{\includegraphics[width=3.8in]{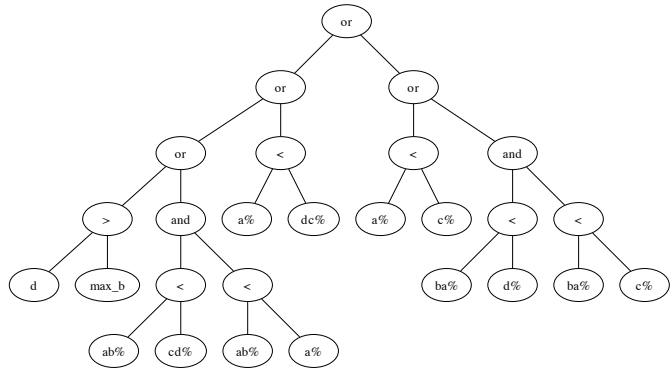}}
\vspace{-1.7cm}
\caption{The ``M`` classification rule.} 
\label{m-rule}
\end{figure}

\begin{figure}
\vspace{-0.1cm}
\centerline{\includegraphics[width=3.8in]{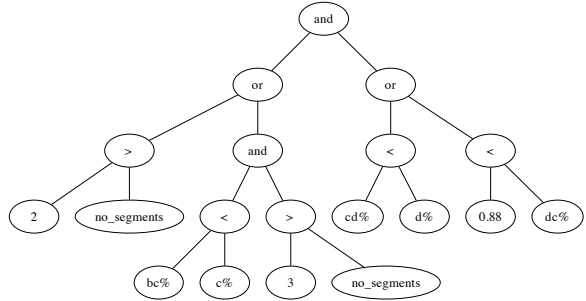}}
\vspace{-2.5cm}
\caption{The ``OB`` classification rule.} 
\label{ob-rule}
\end{figure}

\begin{figure}
\vspace{-3.2cm}
\centerline{\includegraphics[width=3.8in]{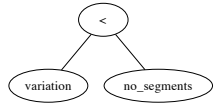}}
\vspace{-3.0cm}
\caption{The ``MTC`` classification rule.} 
\label{mtc-rule}
\end{figure}

Figure \ref{a-rule} gives the A-rule tree in the best team. Its interpretation is straight-forward: symbol $a$ (which represents the smallest Vsh value) has to dominate the segments in the depositional unit. This model accurately describes the composition of channel-axis deposition: massive sandstone without shale (see Section \ref{interpretation}).  The other 4 rules in the best team are given in Figures \ref{oa-rule}, \ref{m-rule}, \ref{ob-rule} and \ref{mtc-rule}. Note that the symbols in the classification rules are shorthand for the symbol thickness. For example, \texttt{a} is the shorthand for \texttt{a\_thickness};  \texttt{cd} is the shorthand for \texttt{cd\_thickness}.

Not all 33 attributes in Table \ref{attributes} are used in these rules to classify deposition types. For example, the thickness of symbol \texttt{b} does not appear in any of the rules. However, this does not mean the classification decision is independent of the presence of symbol "b" in the Vsh data. This is because the presence/absence of symbol \texttt{b} has impact on the attributes database values. For example, the increases of thickness of symbol "b" decreases the percentages of other symbol thickness (\texttt{a\%}, \texttt{c\%}). Meanwhile, the \texttt{total\_thickness} is increased and the number of segments (\texttt{no\_segments}) is increased. Since these values are used by the the classification rule to make classification decision, symbol \texttt{b} has influence on the classification results.

Similarly, symbol ``b`` influences the FST interpretation in the following ways.
\begin{itemize}
\item The symbol is used by the FST to select one of the 5 rules to fire;
\item The symbol is used to decide the next state of FST, which has impact on the next classification rule to fire after reading the next Vsh symbol.
\end{itemize}

Note that when applying these 5 classifiers to a new GR well log, we have to make sure that the well log is from a reservoir which has similar lithology as that of the reservoir from which the GR training data was obtained. Otherwise, these classifiers would not work well.


\section{Evolving Finite State Transducers}\label{fuzzyfst}
With the establishment of the 5 classifiers, we can proceed to the last step (Step 7) of FSTs training.
This FST translates a sequence of Vsh symbols to a sequence of depositional labels.

There are various ways to represent a FST for evolutionary learning, such as the pioneer work by \cite{fogel_etal} and extended work by \cite{fogel}. In this research, we followed the work of \cite{lucas_reynolds} and used two tables to represent a FST. The first table is transition table (see Figure \ref{transition}), which gives the next state of the FST based on the current input symbol and the current state. The second table is output table (see Figure \ref{output}), which gives the output symbol of the FST also based on  the current input symbol and the current state. In our case, the input symbols are the 10 Vsh symbols and the output symbols are the names of the 5 classification rules. 

After a classification rule is decided by the FST, it is applied to the current attribute database. If it returns \texttt{true}, the name of the rule is the output depositional label. Otherwise, the output label is \texttt{null}. 
Consequently, the length of the output deposition labels (e.g. \texttt{A, OA, M, OB, MTC or null}) generated by the FST is always the same as the length of the input Vsh symbols.
The following sub-section describes the workflow of this FST.

\subsection{FST Operating Procedures}
\begin{figure}[htp]
\vspace{-4.5cm}
\centerline{\includegraphics[width=4.5in]{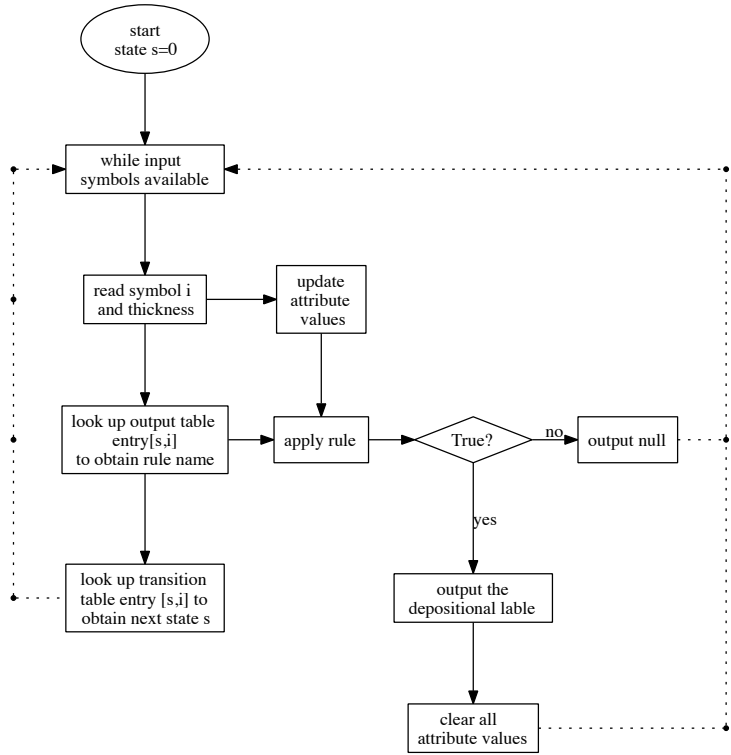}}
\vspace{-2.0cm}
\caption{The operating procedure of the FST.}
\label{FSTflow}
\end{figure}

As shown in Figure \ref{FSTflow}, the FST is initially at state 0 (\texttt{S0}). At each step, one Vsh symbol and the thickness of the segment the symbol represents are processed. The thickness is  used to update the attribute values in the database listed in Table \ref{attributes}. For example, the number of segments (\texttt{no\_segments}) is increased by 1 and the \texttt{total\_thickness} is increased by the symbol's thickness. Meanwhile, based on the current state and the Vsh  symbol, a classification rule in the output table is selected to fire. The name of the selected rule is the proposed deposition type for the sequence of segments that have been processed so far and assigned with \texttt{null} labels. 

After the selected rule is applied to the updated attribute database, it returns a value of \texttt{true} or \texttt{false}. If the value is \texttt{true}, it indicates that the sequence of segments and their associated thickness satisfy the criteria of the proposed deposition type. The depositional label associated with the classification rule is assigned to the current Vsh symbol. Once this happens, all attributes values in the database are set to 0. The system is ready to process the next Vsh symbol and decides what the next depositional label is.

If the classification rule returns \texttt{false}, it indicates that the sequence of segments and their associated thickness do not satisfy the criteria of the proposed deposition type. 
The \texttt{null} label is assigned to the current Vsh symbol.  The current attribute database is retained as it reflects the thickness information of all processed Vsh symbols that have been assigned with  \texttt{null} labels. The FST is ready to process the next Vsh symbol and utilizes the accumulated information in the attribute database to decide what the next depositional label is.

Regardless of the rule firing result, the FST moves to a new state.
The name of the next state is given by the transition table based on the Vsh symbol and the current state.
After moving to the new state, the FST continues processing the next Vsh symbol and repeats the operation until all Vsh symbols are exhausted.

Using the transition table in Table \ref{transition} and the output table in Table \ref{output}, we show the FST operation on input sequence: $(d, 96),(d, 3),(cb, 12.5)$.

With the initial state of \texttt{S0} and the Vsh symbol $d$, the proposed classification rule is OB, according to the output table. 
After updating the attribute database with the thickness of 96, $OB$-rule is applied to the database. Assume the rule returns \texttt{true}, $OB$ becomes the depositional label assigned to symbol $d$.  After that, all attribute values in the database are cleared (become 0). The FST then moves to next state (\texttt{S18}) according to the transition table.

Next, symbol $d$ is processed and the attribute database is updated with the thickness information (3). The proposed rule, according to the output table, is $OA$.
Assume the rule returns \texttt{false} on the updated database, $null$ is assigned to symbol $d$. The FST moves to the next state (\texttt{S10}) without clearing the attribute database. 

The next symbol to process is $cb$. After the attributes database is updated with the thickness of 12.5, it reflects the characteristics of two segments ($d$ and $cb$).
The proposed classification rule, according to the output table, is $A$. Assume the rule returns \texttt{true} on the current attribute database, depositional label $A$ is assigned to symbol $cb$. 
The FST clears the attributes database and moves to next state (\texttt{S7}).
Since there is no more Vsh symbol, the interpretation process terminates.  The depositional labels sequence produced by the FST on the given Vsh symbols is ($OB, null, A$).

\subsection{The Evolutionary System}

The two tables in the FST have the same size of $N_Q \times N_I$, where $N_Q$ is the number of states and $N_I$ is the number of input symbols. 
In this work, $N_I$ is 10: \texttt{a\ldots d} and $N_O$ is 5: \texttt{A, OA, M, OB, MTC}. Based on our estimation of the task complexity, we chose $N_Q$ = 20, which seems to be reasonable or the FST to carry out the inputs/outputs translation. This decision is ad hoc and requires more research to justify its appropriateness (see Section \ref{discussions} more more discussion). 

Based on these parameter values, the number of possible FSTs is:

\begin{equation}
S = N_Q^{N_Q\times N_I} \times N_O^{N_Q \times N_I} \approx 10^{400}
\label{space}
\end{equation}

This is a huge search space and stochastic search methods such as evolutionary algorithms are good candidates to locate a decent solution within a reasonable time frame.
We therefore implemented a GA in Java to train the two FST tables \cite{yu_etal}. 

Unlike the traditional GA \cite{holland}, which evolves vector representation, this GA evolves 2-dimensional tables.
We therefore have to modify the genetic operators to work with the table representation.

This GA only uses mutation operator, which works the same way as that in \cite{lucas_reynolds}. First, a decision is made with equal probability to either mutate the transition table or the output table. A random location is then selected in the chosen table, and the entry there is modified. This ensures that mutation causes at least one change in one of the two tables. 

After that, an iteration is performed over all the table entries apart from the entry just modified, changing each entry with a probability of $\frac{1}{N_Q \times N_I}$. When modifying an entry, a symbol (state name or rule name, depending on the modified table) is chosen uniformly from all possible symbols except the current one. According to \cite{lucas_reynolds}, a single call to the mutation operator is most likely to produce one or two changes to the two FST tables, but can also produce more. The probability that this mutation is applied to the population is 2\%.

The fitness of a FST is calculated based on the depositional labels it produces. After processing a Vsh symbol, a FST produces a depositional label or $null$. The length of the produced labels, hence, is always the same as the length of the Vsh symbols.

We first aligned the 82 Vsh symbols with the 50 depositional labels produced by the stratigrapher, based on the positions where the depositional labels were given.
We then inserted \texttt{null} among the 50 labels on Vsh symbol positions where the stratigrapher did not give a label.
With that, the length of the original 50 depositional labels produced by the stratigrapher is increased to 82.

Next, we aligned this label sequence with that produced by the FST and
the number of mis-match between the two is the FST's fitness. A FST that produce all 50 deposition labels correctly at the correct segment position has fitness value 0. 
\begin{center}
\begin{tabular*}{0.4\textwidth}{@{\extracolsep{\fill}}l p{6cm}}
$fitness_{FST} =  \displaystyle \sum_{i = 1}^{N} mis match_i,$ & N=82.
\end{tabular*}
\end{center}

To encourage FSTs to produce less serious mis-match,  we used the same distance measure used to co-evolve classification team as the secondary criterion for selection (see Section \ref{distance}). If two FSTs have the same number of mis-match, the one with a smaller distance error is the winner during selection.
Table \ref{paraFST} lists the parameter values used to conduct 100 experimental runs.

\begin{table}[h]
\caption{GA parameters for experimental runs.}
\begin{center}
\scalebox{0.88}{
\begin{tabular}{|c|c|c|c|}
\hline
\multicolumn{1}{|c|}{parameter}
& \multicolumn{1}{|c|}{value}
& \multicolumn{1}{|c|}{parameter}
& \multicolumn{1}{|c|}{value} 
\\
\hline
 \hline
tournament selection & size 2 &  elitism & size 1 \\ \hline
population size  & 50&  $N_Q$ & 20 \\ \hline
max generation & 1000 &  mutation rate & 2\% \\ \hline
\end{tabular}
}
\label{paraFST}
\end{center}
\end{table}
\subsection{Results}
\begin{figure}[htp]
\centerline{\includegraphics[width=3.5in]{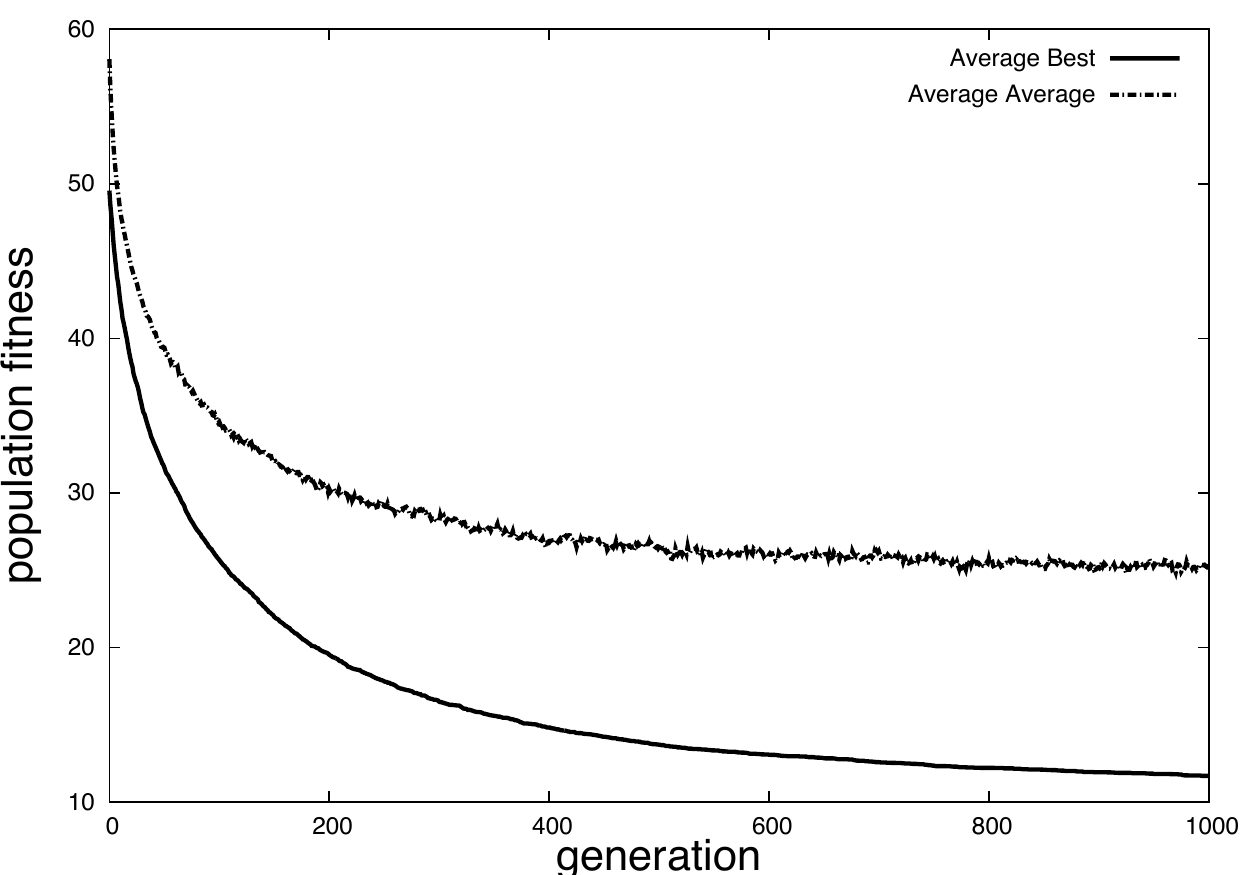}}
\caption{Average and best population fitness.} 
\label{fst}
\end{figure}

Figure \ref{fst} gives the average and the best fitness of the population averaged over 100 runs.
It shows that the populations improved consistently throughout the run of 1,000 generations.
The best FST gives 7 mis-matches to the 82 depositional label produced by the stratigrapher:
\begin{itemize}
\item two $M$ were mis-labeled as $OA$;
\item one $A$ was mis-labeled as $OA$;
\item two $OB$ were labeled as $null$. The segments were then combined with the consecutive segments and labeled as $M$, which was the correct label;
\item two $OB$ were mis-labeled as $M$.
\end{itemize} 

As mentioned in Section \ref{distance}, A and OA have similar lithology while M and OB are close with their depositional ingredients.
The FST translation results are therefore comparable to that produced by the stratigrapher on this data set. 
We list this best transition table and output table in Table \ref{transition} and Table \ref{output}.

\begin{table}[h]
\caption{The best finite state transducer transition table.}
\begin{center}
\tiny
\scalebox{1.3}{
\begin{tabular}{|c||c|c|c|c|c|c|c|c|c|c|}
\hline
\multicolumn{1}{|c|}{input}
& \multicolumn{1}{|c|}{a}
& \multicolumn{1}{|c|}{ab}
& \multicolumn{1}{|c|}{ba}
& \multicolumn{1}{|c|}{b} 
& \multicolumn{1}{|c|}{bc} 
& \multicolumn{1}{|c|}{cb} 
& \multicolumn{1}{|c|}{c} 
& \multicolumn{1}{|c|}{cd} 
& \multicolumn{1}{|c|}{dc} 
& \multicolumn{1}{|c|}{d} 
\\
\hline
\hline
S0&S8&S2&S19&S9&S1&S14&S11&S7&S18&S18\\ \hline	
S1&S9&S17&S4&S5&S3&S2&S14&S12&S2&S10\\ \hline
S2	&S9	&S18	&S1	&S10	&S3	&S9	&S16	&S4	&S1	&S3\\ \hline
S3	&S15	&S9	&S15	&S0	&S16	&S13	&S14	&S17	&S16	&S2\\ \hline
S4	&S0	&S0	&S17	&S8	&S7	&S9	&S3	&S6	&S6	&S13\\ \hline
S5	&S14	&S12	&S9	&S0	&S14	&S16	&S6	&S3	&S3	&S8\\ \hline
S6	&S1	&S14	&S12	&S19	&S3	&S1	&S16	&S1	&S3	&S13\\ \hline
S7	&S17	&S19	&S4	&S19	&S3	&S10	&S6	&S5	&S15	&S15\\ \hline
S8	&S12	&S6	&S5	&S13	&S16	&S1	&S4	&S14	&S16	&S3\\ \hline
S9	&S3	&S19	&S4	&S19	&S11	&S1	&S2	&S15	&S16	&S8\\ \hline
S10	&S7	&S9	&S19	&S6	&S16	&S7	&S11	&S15	&S7	&S6\\ \hline
S11	&S4	&S13	&S19	&S18	&S10	&S8	&S19	&S15	&S2	&S12\\ \hline
S12	&S19	&S1	&S6	&S14	&S11	&S9	&S3	&S18	&S3	&S10\\ \hline
S13	&S10	&S11	&S10	&S11	&S7	&S8	&S3	&S15	&S17	&S6\\ \hline
S14	&S9	&S16	&S0	&S3	&S4	&S3	&S8	&S15	&S5	&S3\\ \hline
S15	&S13	&S13	&S3	&S6	&S9	&S8	&S3	&S7	&S18	&S6\\ \hline
S16	&S18	&S6	&S2	&S5	&S0	&S14	&S10	&S14	&S11	&S4\\ \hline
S17	&S9	&S16	&S4	&S6	&S7	&S6	&S13	&S7	&S9	&S4\\ \hline
S18	&S1	&S12	&S19	&S6	&S2	&S9	&S0	&S0	&S5	&S10\\ \hline
S19	&S13	&S2	&S15	&S18	&S14	&S0	&S18	&S2	&S12	&S0\\ \hline
\end{tabular}
}
\label{transition}
\end{center}
\end{table}

\begin{table}[h]
\caption{The best finite state transducer output table.}
\begin{center}
\tiny
\scalebox{1.1}{
\begin{tabular}{|c||c|c|c|c|c|c|c|c|c|c|}
\hline
\multicolumn{1}{|c|}{input}
& \multicolumn{1}{|c|}{a}
& \multicolumn{1}{|c|}{ab}
& \multicolumn{1}{|c|}{ba}
& \multicolumn{1}{|c|}{b} 
& \multicolumn{1}{|c|}{bc} 
& \multicolumn{1}{|c|}{cb} 
& \multicolumn{1}{|c|}{c} 
& \multicolumn{1}{|c|}{cd} 
& \multicolumn{1}{|c|}{dc} 
& \multicolumn{1}{|c|}{d} 
\\
\hline
\hline								
S0&OA	&OB	&OB	&OB	&OA	&A	&MTC	&M	&OB	&OB\\ \hline
S1&OA	&MTC	&OB	&MTC	&MTC	&M	&M	&M	&OB	&M\\ \hline
S2&OB	&OA	&MTC	&OA	&MTC	&M	&OB	&M	&OA	&A\\ \hline
S3&M	&OA	&OA	&A	&M	&MTC	&OB	&M	&OB	&OB\\ \hline
S4&M	&MTC	&A	&OB	&M	&OB	&OA	&OB	&MTC	&OB\\ \hline
S5&M	&A	&M	&OA	&M	&OA	&A	&A	&MTC	&A\\ \hline
S6&A	&M	&OA	&MTC	&MTC	&OA	&OB	&OA	&M	&A\\ \hline
S7&OA	&M	&M	&OB	&M	&A	&M	&OA	&MTC	&M\\ \hline
S8&OA	&OB	&MTC	&MTC	&OA	&OB	&MTC	&MTC	&M	&M\\ \hline
S9&OA	&OA	&A	&A	&MTC	&MTC	&MTC	&M	&OA	&MTC\\ \hline
S10&OA	&A	&OA	&OA	&OB	&A	&MTC	&OA	&A	&OA\\ \hline
S11&MTC	&OB	&OB	&OB	&A	&MTC	&M	&A	&MTC	&OA\\ \hline
S12&M	&M	&M	&MTC	&OA	&A	&OB	&MTC	&MTC	&OB\\ \hline
S13&OA	&MTC	&M	&M	&OA	&MTC	&OB	&OB	&OA	&OA\\ \hline
S14&A	&MTC	&MTC	&MTC	&MTC	&OA	&OB	&M	&MTC	&M\\ \hline
S15&M	&OB	&OB	&A	&MTC	&A	&OB	&MTC	&A	&OB\\ \hline
S16&OA	&MTC	&A	&MTC	&MTC	&OB	&M	&MTC	&OA	&OA\\ \hline
S17&A	&OA	&A	&OA	&M	&OA	&OB	&OB	&M	&M\\ \hline
S18&A	&MTC	&OB	&OA	&MTC	&A	&OB	&MTC	&OB	&OA\\ \hline
S19&A	&MTC	&A	&A	&OA	&OB	&MTC	&A	&OA	&MTC\\ \hline
\end{tabular}
}
\label{output}
\end{center}
\end{table}
\normalsize

\section{Limitations and Future Works} \label{discussions}

 Using one GR well log data set, we demonstrated that the developed methodology
 and the designed workflow can produce FST models that interpret deepwater reservoir depositional environments adequately.
 However, the applicability of the method to the deepwater reservoir industry is still unknown until we have tried and test it on several more data sets.
 Additionally,  we need to investigate some practical issues to assure the framework works for a wide range of deepwater reservoirs. 
 
 First,  the data set we used to develop the method is reasonably clean, which helped train good quality models. It is not clear if the method would work well on noisy data.
 The proposed interpretation system is a fuzzy model, which should perform well on noisy and uncertain data.
 However, we have only implemented the crisp version of the  system. To fully realize the fuzzy version of interpreter, the 5 classifiers and the
 FST need to manipulate fuzzy symbols. For example, we need to provide fuzzy definition for the 33 attributes and the 6 Boolean operators. 
 Similarly, the FTS should decide the next state and the classification rule to fire based on the fuzzy Vsh symbol. 
 We plan to develop the fuzzy version of the interpretation system in our future work.

Second, with 6,150 data points, the computation time to develop the interpretation system manageable.
It took a single-core Linux machine seconds to complete data segmentation and symbol mapping (Step 1--4).
The machine took 2 hours to train 5 classifiers (Step 6) and 2 hours to train the FST (Step 7). 
We are not clear if the method would work with more or less amount of data. 
Our hypothesis is that if we apply the method to multiple GR logs from more than one reservoirs at the same time, the resulting models should be better and more robust.
We will test this hypothesis in our future works.

There are many areas in the proposed framework that we can improve.
First, the proposed compromised approach to decide the number of segments may not  be optimal.
Although the number of segments could not be too large due to the computational complexity, the capacity of capturing the data trend is also important.  
In our future work, we will assign weight factors to the two objectives of \texttt{error} and \texttt{the number of segments} in Equation (3).
 We will evaluate the trade-off by adjusting the weights to segment the well logs to identify a good balance.

 
 
 

Second, the FST is a one-pass algorithm without back track. After reading a Vsh symbol, 
the FST gives the depositional label based on the selected classification rule and the current 
attribute database.
This interpretation decision can not be revised later.
It might be interesting to have a back track FST that can revise previous interpretation if it finds a better interpretation.
We will research this alternative FSM in our future work.

Next, the FST performs one look-ahead of the new Vsh symbol to make interpretation decision. One-look-ahead is the simplest FST. It might be interesting to extend the FST to look ahead more than one Vsh symbol at a time to make interpretation decision. We will investigate this alternative FSM in our future work.

Last, the FST tables show that certain sets of inputs and certain sets of states have similar behavior.
This suggests that there might be room for aggregation by reducing the input size ($N_I$) and state size ($N_Q$) of the FST.
In our future work, we will investigate the granularity of the FST. 
 
\section{Concluding Remarks}\label{conclude}
Reservoir characterization continues to present new challenges as exploration moves to new territories, such as
deepwater. We have developed a stratigraphic interpretation framework that analyzes depositional information to 
improve reservoir characterization. In this framework, one critical step is the interpretation of the stratigraphic 
components in the reservoir. The quality of the interpretation can impact the prediction of oil recovery of the reservoir.

This research developed a novel method and designed a workflow to automate this interpretation process.
In particular, we treated the interpretation task as a language translation problem and applied various computational intelligence techniques
to train FST models that carries out the interpretation task. 
To the best of our knowledge, this is the first time the language-translation approach with a FST model is used to interpret well log data.
We also showed the interpretation resulst produced  by the FST are comparable to that produced by the stratigrapher on this data set. 

With this encouraging initial result, we will continue our future work on the areas outlined in Section \ref{discussions}.
We will also help stratigraphers to understand the FST model training process to gain their acceptance of the developed models.

\end{document}